\documentclass[twoside]{article}

\usepackage{graphicx}
\usepackage{amsfonts}
\usepackage[table]{xcolor}
\definecolor{LightCyan}{rgb}{0.88,1,1}
\usepackage[accepted]{aistats2021}
\def\PRsub#1#2{p_{#1}(\,#2\,)}
\def\PR#1{\PRsub{}{#1}} % p(\, #1\,)

\def\CPR#1#2{\PRsub{}{#1\,|\,#2}}

\def\ie{{\em i.e.},\ }
\def\eg{{\em e.g.},\ }

\begin{document}

% \runningtitle{Probabilistic Labels for Image-based Classifiers}

\runningauthor{Vega, Gorji, Zhang, Qin, Rakkunedeth, Kapur, Jaremko, Greiner}

\twocolumn[

\aistatstitle{Sample Efficient Learning of Image-Based Diagnostic Classifiers Using Probabilistic Labels}

\aistatsauthor{Roberto Vega$^{1,2}$ \And Pouneh Gorji$^{1,2}$ \And  Zichen Zhang$^{1,2}$ \And Xuebin Qin$^{3}$}
\aistatsauthor{Abhilash Rakkunedeth Hareendranathan$^{1}$ \And Jeevesh Kapur$^{3,4}$}
\aistatsauthor{Jacob L. Jaremko$^{1,3}$ \And Russell Greiner$^{1,2}$}

\aistatsaddress{ \\$^{1}$University of Alberta\\Canada\\ \{rvega, rgreiner\}@ualberta.ca \And  \\$^{2}$Alberta Machine\\Intelligence Institute, Canada\\ \And \\$^{3}$Medo.AI\\Canada \And \\$^{4}$National University of \\Singapore, Singapore}
]

\begin{abstract}
  Deep learning approaches often require huge datasets to achieve good generalization. This complicates its use in tasks like image-based medical diagnosis, where the small training datasets are usually insufficient to learn appropriate data representations. For such sensitive tasks it is also important to provide the confidence in the predictions. Here, we propose a way to learn and use {\em probabilistic labels} to train accurate and calibrated deep networks from relatively small datasets. We observe gains of up to 22\% in the accuracy of models trained with these labels, as compared with traditional approaches, in three classification tasks: diagnosis of hip dysplasia, fatty liver, and glaucoma. The outputs of models trained with probabilistic labels are \emph{calibrated}, allowing the interpretation of its predictions as proper probabilities. We anticipate this approach will apply to other tasks where few training instances are available and expert knowledge can be encoded as probabilities. \end{abstract}

\section{INTRODUCTION}

Deep learning has proven useful in solving problems that involve structured inputs, such as images, voice, and video~\cite{goodfellow2016deep}. Those inputs tend to be high-dimensional, so machine learning models use many parameters (in the order of $10^4-10^7$) to provide accurate predictions. Fitting such large models usually requires datasets whose number of instances are at least in the same range as the number of parameters, complicating its use in many fields.

Areas like medical imaging can benefit from machine learning, since it might reduce the time that radiologists spend manually analyzing images~\cite{pesapane2018artificial}. Machine learning algorithms aim to learn relevant patterns in the data, and then use these patterns to make accurate diagnostic and prognostic predictions. However, imaging-based medical diagnosis has some characteristics that makes it different from traditional image classification tasks: 
(1) A small number of labeled instances available for training, usually in the range from few hundreds to few thousands. 
(2) Uncertainty in the labels used as ground truth, since different medical experts often give different labels to the same image. 
(3) Importance of providing not only the predictions, but also the degree of confidence in them. 
(4) A vast medical literature that can be exploited to learn the models with less training instances.

These four characteristics suggest it is important to find learning methodologies that are sample efficient, allow the incorporation of prior knowledge, and produce accurate and calibrated models. We argue that traditional learning paradigms in deep learning do not meet these characteristics, so we propose the use of {\em probabilistic labels}\ as an alternative way of achieving those objectives. The basic idea is: instead of using discrete labels as the target during the training procedure, use dense vectors whose \emph{k}-th entry represents the probability of belonging to the \emph{k}-th class. We anticipate that encoding the targets as real numbers, instead of just categorical data, might compensate for the small number of training instances by providing more information per instance in the form of those probabilistic labels.

A key contribution of our approach is a mechanism to compute the probabilistic labels when we only have access to the discrete ones. We propose encoding the medical knowledge as probabilities. First, during the training phase, we extract from every image a set of features considered in the medical literature as relevant for making a diagnosis. Then, we build a probabilistic model over these features and the categorical labels. Next, for every training instance, we compute the probability of each class given the features. Finally, we use these probabilities as targets of a convolutional neural network whose inputs are the raw images. This approach is similar to model distillation~\cite{hinton2015distilling}, but instead of distilling a complex model, we 'distill the medical knowledge'.

To help motivate this work, below we discuss three natural questions: why use deep learning if the relevant features are known?, what if the probabilistic labels are misleading?, and is the extraction of the features worth the effort?

Even when medical experts can identify the relevant markers in the image (\eg an important statistic about a shape -- see Fig~\ref{fig:hipdysplasia}), they still have to compute the features manually. Since these features are mostly visual, or require special software to perform measurements, it is unclear how to automate this process during inference. By combining deep learning with probabilistic labels, the network learns a good representation of the data using a limited amount of training instances. This representation removes the need of any manual computation of features during the inference process. In other words, a deep learning model can provide a prediction directly from the image, without the need of any human intervention.

We also incorporated a regularization parameter, $\lambda$, that controls the influence of the probabilistic labels during training. These labels help the algorithm to learn the model parameters accurately with fewer training instances, assuming the labels are a good approximation of the real probabilities. Even with misleading probabilistic labels, our approach can recover the true probabilities from the discrete labels, given a large enough dataset.

Our experiments suggest that the effort of providing the extra features during training is justified. We used first a simple toy dataset that exemplifies the advantages of training with probabilistic labels. Then, we used 3 real-world imaging datasets for the diagnosis of (respectively) hip dysplasia, glaucoma, and fatty liver. The use of probabilistic labels not only improved the classification accuracy up to 22\%, relative to the use of discrete labels, but it also produced models that are calibrated -- \ie the output of the model can be interpreted as probabilities~\cite{guo2017calibration,hosmer1980goodness}.

Section~\ref{sec:foundations} describes relevant literature for the problem of calibration and sample efficient learning. Section~\ref{sec:prob_labels} describes probabilistic labels and justifies their use to train deep learning models. Section~\ref{sec:experiments} compares the performance of models trained using probabilistic labels versus other training approaches. Finally, Section~\ref{sec:discussion} highlights the important elements of this approach, emphasizing where it is expected to excel.

\section{FOUNDATIONS AND RELATED WORK}\label{sec:foundations}

The objective of supervised learning for classification problems is to find a mapping $f : X \to Y$; $X \in \mathbb{R}^p$, $Y \in \{1,2,...,K\}$, that minimizes a cost, $c(y, f(x))$, between the true label $y$ and the predicted label $\hat{y} = f(x)$. Each $x \in X$ is a \emph{p}-dimensional vector containing the values of a set of features, $y \in Y$ indicates the categorical class to which $x$ belongs, and $K$ is the number of classes.

\subsection{Soft Labels}
Traditionally, the label (\emph{y}) of each instance is encoded as a one-hot vector -- \eg the encoding $h_i = [0, 1, 0]$ indicates that $x_i$ belongs to the second class, $y_i = 2$, since the 1 is located at the 2\emph{nd} entry. We call this a \emph{hard label}. Note that hard labels allocate all the probability mass to a single class, encouraging big differences between the largest logit and all others in networks that use the soft-max activation function in the output layer~\cite{szegedy2016rethinking}. This is undesirable in applications like medical imaging, where many cases are ``bordeline'', in that it is not clear to which class they belong. Creating an artificial gap between logits might then cause overfitting and reduce the ability of the network to adapt~\cite{szegedy2016rethinking}.

By contrast, {\em soft labels} encode the label of each instance as a vector of real values,
whose $k$-th entry represents $p(Y = k\ |\ X=x) \in [0,1]$~\cite{geng2016label}. For example, the soft-label $s_i = [0.1, 0.7, 0.2]$ indicates that $p(Y = 2\ |\ X=x_i) = 0.7$. By using real numbers instead of single bits, soft-labels provide to the learning algorithm extra information that often reduces the number of instances required to train a model~\cite{hinton2015distilling}, while improving the performance during inference~\cite{gao2017deep,geng2016label,imani2018improving}.

The main challenge in using soft labels is their proper computation. Nguyen et al. (2011) proposed directly asking domain experts for their best estimates of $p(Y = k\ |\ X=x)$. Models trained with these soft labels learned more accurate classifiers, using fewer labeled instances, than classifiers trained with hard labels~\cite{nguyen2011learning}. One complication is that human experts struggle to give reliable and consistent estimates of the probabilities. One effective way of reducing this problem is to group the probabilities into bins~\cite{xue2017efficient}. However, this still relies on human estimates.

A different approach is the use a smoothing parameter that distributes a fraction of the probability mass, $\epsilon$, over all the possible classes (\eg if $\epsilon = 0.1$, then the label $[0,1]$ becomes $[0.1, 0.9]$). This solution achieved an increase of 0.2\% in accuracy on the ImageNet dataset~\cite{szegedy2016rethinking}. Pereira et al. (2017) suggested a similar approach: Directly penalize `confident predictions' of a neural network by adding the negative entropy of the output to the negative log-likelihood during training~\cite{pereyra2017regularizing}. This strategy, whose performance was similar to label smoothing, penalizes the allocation of all the probability mass on a single class at inference time. Note that these approaches apply the same smoothing to all the labels; however, Norouzi et al. (2016) empirically demonstrated that not all the classes should receive the same probability mass~\cite{norouzi2016reward}. In fact, one can argue that arbitrarily penalizing confident predictions is not a good strategy in the medical domain, since there are cases when we want the classifier to have high confidence in the predictions. 

A third strategy, which is based on work on model compression~\cite{ba2014deep,bucilua2006model}, is to use model distillation~\cite{hinton2015distilling, HintonSmoothing}. Here, we first train a complex model that outputs a vector of real numbers, whose \emph{k}-th entry is interpreted as the probability of belonging to the \emph{k}-th class. Then, train a second, simpler, model whose target is the output of the first model. This is a very effective approach, but it requires enough data to train that accurate complex model first. Unfortunately, for medical tasks, the scarcity of labeled data complicates the use of this solution.

The results obtained by the aforementioned approaches strongly argue for the use of soft labels in classification tasks, but they highlight two unresolved problems: (1) It is still not clear how to properly obtain the labels, and (2) these approaches ignore the original true labels, so unreliable soft-labels will lead to unreliable results. We propose the use of \emph{probabilistic labels} to alleviate those problems. First, train a simple probabilistic model based on the hard-labels, whose features are manually extracted by medical experts. The predictions of this model, which encode the expert medical knowledge, can be treated as reliable and consistent soft labels. Next, ``distill'' the knowledge of this probabilistic model and transfer it to a deep neural network. This neural network will receive as an input a raw image, and uses, as targets both, the soft-labels produced by the probabilistic model, and the original hard labels. The influence of each label is determined by a regularization parameter $\lambda$, which can be determined using cross-validation.

\subsection{Calibration}
A probabilistic classifier is considered ``calibrated'' if the probabilities it returns are good estimates of the actual likelihood of an event~\cite{guo2017calibration,haider2018effective,hosmer1980goodness}. For example, if a calibrated classifier predicts that the probability of having a disease is 30\% for 10 individuals, then we would expect 3 of those individuals to actually have the disease. Calibration is particularly relevant for critical decision-making tasks. Common metrics to determine if a predictor with parameters $\theta$, $p_{\theta}(Y\ |\ X)$, outputs calibrated probabilities is the Hosmer-Lemeshow goodness-of-fit statistical test~\cite{hosmer1980goodness} and the expected calibration error~\cite{guo2017calibration}. 

Typically, the output of a neural network that uses a sigmoid or soft-max activation function in its last layer is interpreted as the probability of the classes given the input~\cite{goodfellow2016deep}. These activation functions indeed output values in $[0, 1]$ whose values add to 1; however, there is evidence that traditional learning approaches in modern neural networks lead to poorly calibrated models and therefore do not represent ``real probabilities"~\cite{guo2017calibration}.

The poor calibration problem becomes evident after analyzing the cross-entropy cost function, which is commonly used to train classifiers:
\begin{equation}
    c\left(y, f(x)\right)\ =\ \frac{1}{M}\sum_{i=1}^M \sum_{k=1}^K p(y_i = k\ | \ x_i)\ \log(f(x_i, k))
\end{equation}

\noindent where \emph{M} is the number of training instances, \emph{K} is the number of classes, and $f(x_i, k)$ is the predicted probability that $x_i$ belongs to the \emph{k}-th class. Note that for a fixed \emph{x} and \emph{k}, the prediction that minimizes the cost, when using hard-labels, is:

\begin{equation}
    f(x, k)\ =\ \frac{1}{M_x}\sum_{i=1}^{M_x} I(y_i = k)
\end{equation}

\ie the optimal prediction is the proportion of observations labeled with class $Y=y$ out of the total number of instances, $M_x$, where $X=x$ . By the law of large numbers, as $M \to \infty$, $f(x, k) \to p(Y = k\ | \ x)$; however, when the number of training instances is small, $f(x, k)$ might not be a good approximation of $p(Y = k\ | \ x)$, meaning the predictions are not calibrated.

A second problem for calibration arises when the inputs are high-dimensional. In gray-scale medical images, most of the pixels take values in the interval $[0, 255]$. Therefore, the sample space is $[0, 255]^p \times \{0, 1, \dots, K\}$, where $p= |x|$ is the number of pixels, which is typically around $10^4$.  The sample space immediately highlights the difficulties that any learner has to learn $p(Y\ |\ X)$: It needs a very large number of instances to approximate this probability directly from the images and the hard labels. 

One way of improving the calibration of traditional machine learning models (linear models, decision trees, etc.) is via Platt scaling or isotonic regression~\cite{niculescu2005predicting}. Similarly, the use of a temperature parameter helps the calibration on modern neural networks~\cite{guo2017calibration}. These simple, yet effective, methods improve the calibration of the predictions. However, none of these methods improve the accuracy of the models --\ie they only modify the \emph{confidence} in the predicted class for a novel instance. Here, we empirically show that, by using \emph{probabilistic labels}, it is possible improve both the calibration of the predictions and the classification performance.

\section{PROBABILISTIC LABELS}\label{sec:prob_labels}
The high dimensionality of images poses important challenges for learning calibrated and accurate predictors~\cite{friedman2001elements}, so dimensionality reduction is a common step in the machine learning pipeline~\cite{guyon2008feature}. Similarly, medical experts do not analyze the images at the pixel level. Instead, they are trained to identify relevant features in the images, and then combine those features to produce the diagnosis. 

The idea behind probabilistic labels is to first obtain the relevant features $,Z(X),$ from raw images $X$. Then, use the hard labels along with a probabilistic model to estimate $p(\ Y\ |\ Z(X)\ )$; see Section~\ref{sec:hip}. Since those features are assumed to be a good representation of the image, then it is valid to assume that $p(Y\ |\ Z(X)) \approx p(Y\ |\ X)$. Since $|Z(X)| \ll |X|$ --\ie $Z(X)$ has fewer features than the raw $X$, we expect the estimation of $p(Y\ |\ Z(X))$ to be more accurate than the one of $p(Y\ |\ X)$, given the same number of training instances. 

The last step is to ``distill'' the medical expert knowledge encoded in the probabilistic model. To do this, train a deep learning model using the raw images $X$ as inputs, and $p(Y\ |\ Z(X)\ )$ as targets. The learning problem is then to learn function $q(x) = \CPR{y}{z(x)}\ \in [0,1]^K$ that maps a medical image, $x$, to the probability of being classified as each of the \emph{K} classes. 

To learn such a function we apply a learning algorithm, $L(\cdot)$
to a labeled training set with $n$ instances $D = \{[x_1,\,\CPR{y_1}{z_1}],\ \dots,\ [x_n,\,\CPR{y_n}{z_n}]\}$
to get an estimate of the function, $\hat{q} = L(D)$.  It is then possible to make predictions on new instances $\hat{y}_{new} = \hat{q}(x_{new})$. Note that once the model has been learned, the only input is a raw image, and it is no longer necessary to compute the feature vector $z$.

Given the success of deep learning models on images, we chose the learning algorithm $L(\cdot)$ to be a convolutional neural network with fully connected layers with a softmax activation function in the last layer. The detailed architecture will be described in the next section. Since this target vector is now a probability distribution, it makes sense to use a loss function that measures the distance between the target distribution and the predicted outputs of the network.

A common measure for distance between probability distributions is the KL divergence:
\begin{equation}
D_{KL}(\,P\, ||\, Q_\theta\,)\ =\ -\sum_{y \in Y}\CPR{y}{x}\, \log \left( \frac{q_\theta (y\ | \ x)}{\CPR{y}{x}}\right)
\end{equation}
\noindent where $\CPR{y}{x}$ is the real conditional distribution of the labels given the inputs, 
and $Q_\theta (y\ |\ x)$ is the probability distribution predicted by a model parameterized by $\theta, q_\theta (\cdot)$. 
Minimizing the KL divergence between 
the % both 
distributions is equivalent to minimizing the negative cross entropy:

\begin{equation}\label{eq:KL_binary}
    \theta^*\ =\ \arg \min_\theta -\sum_{i=1}^m \sum_{k=1}^K p(Y=k | x_i)\  \log \left( q_{\theta}(x_i,k)\right)
\end{equation}

Note that this objective is identical to the one we use for training with hard labels;
the only difference is that instead of using the indicator function as target, $\CPR{Y=k}{x} = I(Y=k | x)$, we use the \emph{probabilistic label} $\CPR{Y=k}{x} = \CPR{Y=k}{z(x)}$.

The quality of the learned model will depend on the quality of the estimation of the probabilistic label. Suppose that for a fixed $x$ the real probability is $p(y=1\ |\ x) = 0.75$, but we only have access to 5 instances (2 positive, 3 negative). Using the traditional approach, the model that optimizes Eq.~\ref{eq:KL_binary} will converge to $f_\theta(x) = 0.4$.
If our guess of the probabilistic label is $y_{pr} = 0.73$, then we can expect a better performance by using the probabilistic label. Note that although the evidence given by the hard labels indicates that $f_\theta(x) = 0.4$, our ``confidence'' in that evidence is small, due to the small number of instances.

The opposite effect can also happen. When the estimation of the probabilistic label is incorrect, the model will converge to that label regardless of the evidence given by the hard labels. Ideally, we should find a balance between the influence of hard and probabilistic labels. We propose to achieve this behavior by training our model in two steps: (1) Let $\theta_{p}$ be the parameters of a model $q_{\theta_{p}}(x)$ trained to optimize Eq.~\ref{eq:KL_binary} using the probabilistic labels  $\CPR{Y}{z_i}$. (2) Use the  weights $\theta_{p}$ as a prior to learn a second model, with parameters $\theta$, that uses the \emph{hard labels} and optimizes the regularized cross-entropy:
\begin{equation}\label{eq:reg_CE}
\begin{gathered}
    \theta^* = \arg \min_\theta -\sum_{i=1}^m \sum_{k=1}^K I(Y=k | x_i)  \log \left( q_{\theta}(x_i,k)\right) \\+ \lambda ||\theta - \theta_{p}||_2^2
`\end{gathered}    
\end{equation}

Intuitively, this loss function penalizes deviations from the model learned with probabilistic labels. Note that as the number of instances, $m$, increases, the influence of the regularization term decreases. It can be shown that this regularized loss function is equivalent to setting a Gaussian prior on the weights~\cite{murphy2012machine}. The mean of this Gaussian prior is $\theta_{p}$, and the covariance matrix is $\frac{1}{2\lambda} I$, where $I$ is the identity matrix. Therefore, a high value of $\lambda$ means that the confidence in the probabilistic labels is high. However, as the number of instances increases, the influence of the prior decreases. In practice, we can set the value of the regularization parameter $\lambda$ using cross-validation. Although we describe this algorithm in two steps, in practice both models are implemented under a single routine.

\subsection{Example: Hip dysplasia}\label{sec:hip}
\begin{figure}[t]
\centering
\includegraphics[width=0.9\columnwidth]{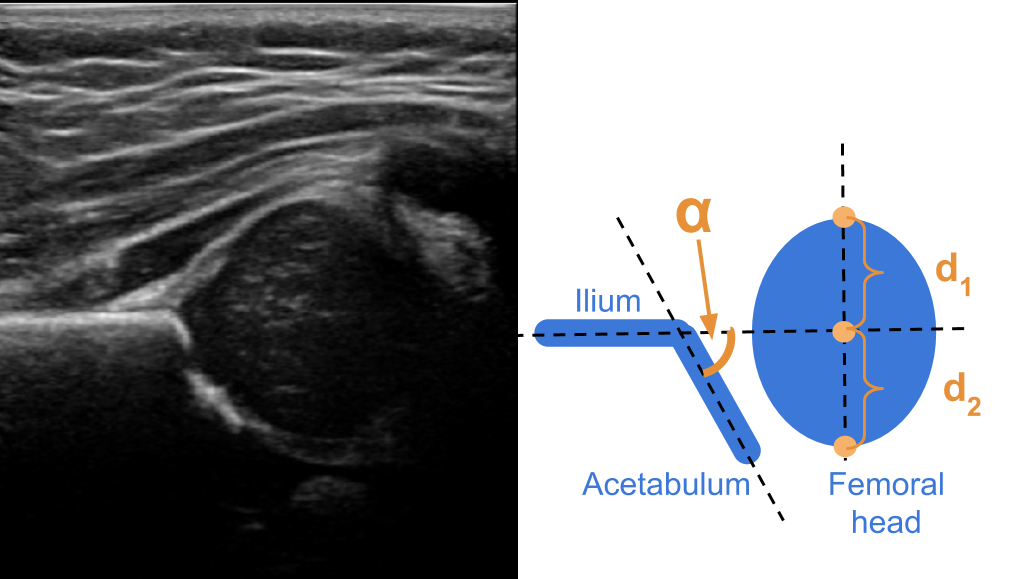} % Reduce the figure size so that it is slightly narrower than the column. Don't use precise values for figure width.This setup will avoid overfull boxes. 
\caption{\textbf{(Left)} Typical 2D ultrasound image of the hip. \textbf{(Right)} The structures of interest: the acetabulum, the ilium and the femoral head,
as well as the angle between the acetabulum and ilium ($\alpha$), and the information for computing the coverage: $c\ =\ \frac{d_2}{d_1 + d_2}$}
\label{fig:hipdysplasia}
\end{figure}

Developmental dysplasia of the hip is a deformity of the hip joint at birth that affects close to 3\% of infants~\cite{hareendranathan2017toward}. Ultrasound imaging is one way to diagnose this condition. To do so, the medical expert measures the angle $\alpha$ between the acetabulum and ilium, and the coverage $c$ (ratio between the two segments $d_1$ and $d_2$); as shown in Figure~\ref{fig:hipdysplasia}~\cite{graf1984fundamentals,harcke2017hip,hareendranathan2017toward}. 

\begin{figure}
\centering
\includegraphics[width=0.9\columnwidth]{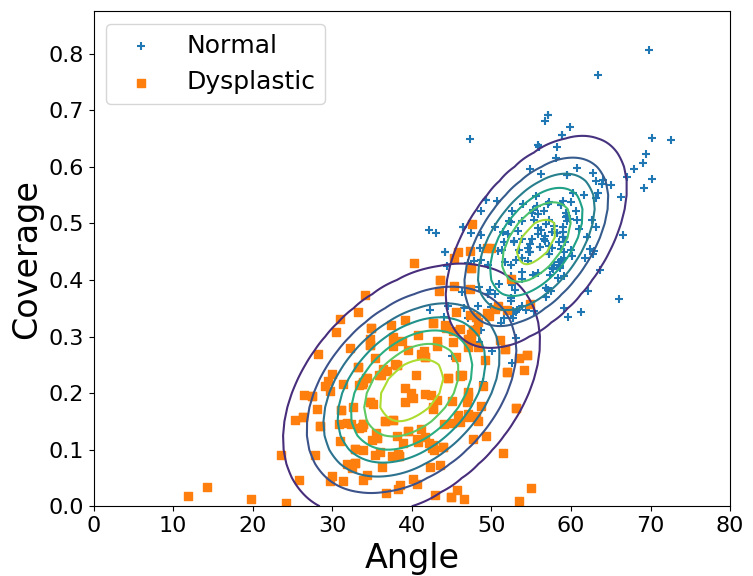} % Reduce the figure size so that it is slightly narrower than the column. Don't use precise values for figure width.This setup will avoid overfull boxes. 
\caption{Gaussian distributions learned for images classified as normal or dysplastic, based on the angle~($\alpha$) and 
coverage~($c$)}
\label{fig:PGM_Dysplasia}
\end{figure}

We can encode this knowledge as a simple probabilistic model where the random variable $Y \in \{0, 1\}$ encodes healthy people as $y=0$ and people with hip dysplasia as $y=1$, 
and the random variable $Z \in [0, \pi] \times \mathbb{R}$ is a vector that contains the pair of computed features: (angle, coverage). The sample space for this new model is $[0, \pi] \times \mathbb{R} \times \{0, 1\}$, where the last bit is the label. Note that this is much simpler than the sample space of the entire image. 

Although the proper probability distributions for modelling angles and ratios are the Von-Mises distribution and a ratio distribution, we assume that a bi-variate Gaussian with full covariance matrix is a good approximation for $P(Z\ |\ Y)$. Figure~\ref{fig:PGM_Dysplasia} shows a scatter plot of the values for $\alpha$ and $c$ for both, people with dysplasia and healthy controls.

The computation of the soft labels $\CPR{Y}{Z=z_i}$ is given by Eq~\ref{eq:bayes}, where $\CPR{Z}{y=i} \sim {\cal N}(\mu_i, \Sigma_i)$, and $\PR{Y=i}$ is the prior probability of having (or not having) dysplasia. It is straightforward to estimate the parameters $\mu_i$, $\Sigma_i$ from the data, while the prior probabilities $\PR{Y=i}$ can be extracted from medical literature. Note that this prior will play a fundamental role in determining if the probabilities are calibrated, as its value should reflect the probability that we expect to observe \emph{where the model will be used}. 

\begin{equation}\label{eq:bayes}
    \CPR{Y=1}{z_i}\ =\ \frac{\CPR{z_i}{Y=1}\,\PR{Y=1}}
    {\sum_{j\in \{0,1\}}\CPR{z_i}{Y=j}\,\PR{Y=j}}
\end{equation}

The last step consists in training a model using the soft labels obtained with Eq.~\ref{eq:bayes}, along with the cost functions described by Eq.~\ref{eq:KL_binary} and Eq.~\ref{eq:reg_CE}. Note that, although we modelled the features as a bi-variate Gaussian, this approach is very general and admits other distributions as well. Probabilistic graphical models offer a good set of tools for modelling distributions with a larger number of features, or that require a combination of discrete and continuous covariates~\cite{hojsgaard2012graphical, koller2009probabilistic,lauritzen1996graphical}.

\section{EXPERIMENTS AND RESULTS}\label{sec:experiments}

% In order to test the hypothesis that probabilistic labels improve the performance (in terms of accuracy and calibration) of deep learning models we used the 4 datasets described below:

\subsection{Experiment 1: Simulated Data} 
\begin{figure}
\centering
\includegraphics[width=0.9\columnwidth]{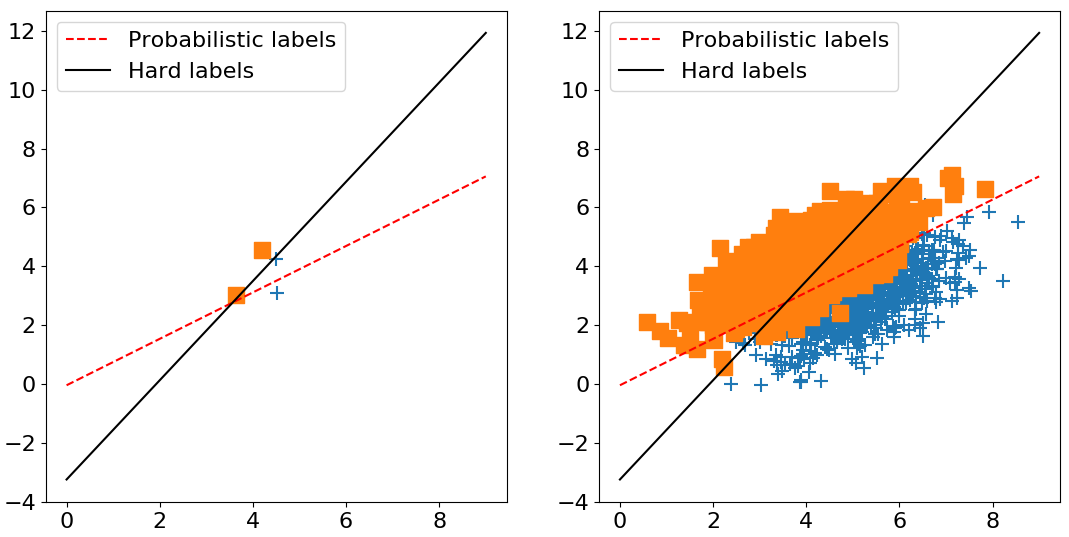} % Reduce the figure size so that it is slightly narrower than the column. Don't use precise values for figure width.This setup will avoid overfull boxes. 
\caption{Decision boundaries learnt with hard labels and soft labels in the training set (left), and how well they generalize to the test set (right).}
\label{fig:soft_hard}
\end{figure}

We generated 2,000 instances from a mixture of 2 bi-variate Gaussians (1,000 from each Gaussian) and used it as a test set. 
The parameters of the Gaussians were: $\mu_1 = \begin{bmatrix}5\\3 \end{bmatrix}$, $\mu_2 = \begin{bmatrix}4\\4 \end{bmatrix}$, $\Sigma_1 = \begin{bmatrix}1 & 0.5\\0.5 &1 \end{bmatrix}$, $\Sigma_2 = \begin{bmatrix}1 & 0.7\\0.7 & 1 \end{bmatrix}$. 
Additionally, we generated 60 instances for training purposes (30 from each Gaussian). The machine learning task was to use the training instances to build a classifier that assigned every instnace to one of the two possible classes: Gaussian 1, or Gaussian 2. Figure~\ref{fig:soft_hard} (right) shows the test instances generated for the experiments.

% \begin{figure}
% \centering
% \includegraphics[width=0.8\columnwidth]{Images/TwoGaussians.png} % Reduce the figure size so that it is slightly narrower than the column. Don't use precise values for figure width.This setup will avoid overfull boxes. 
% \caption{Test set from the simulated data.}
% \label{fig:simulated}
% \end{figure}

\begin{figure}
\centering
\includegraphics[width=0.99\columnwidth]{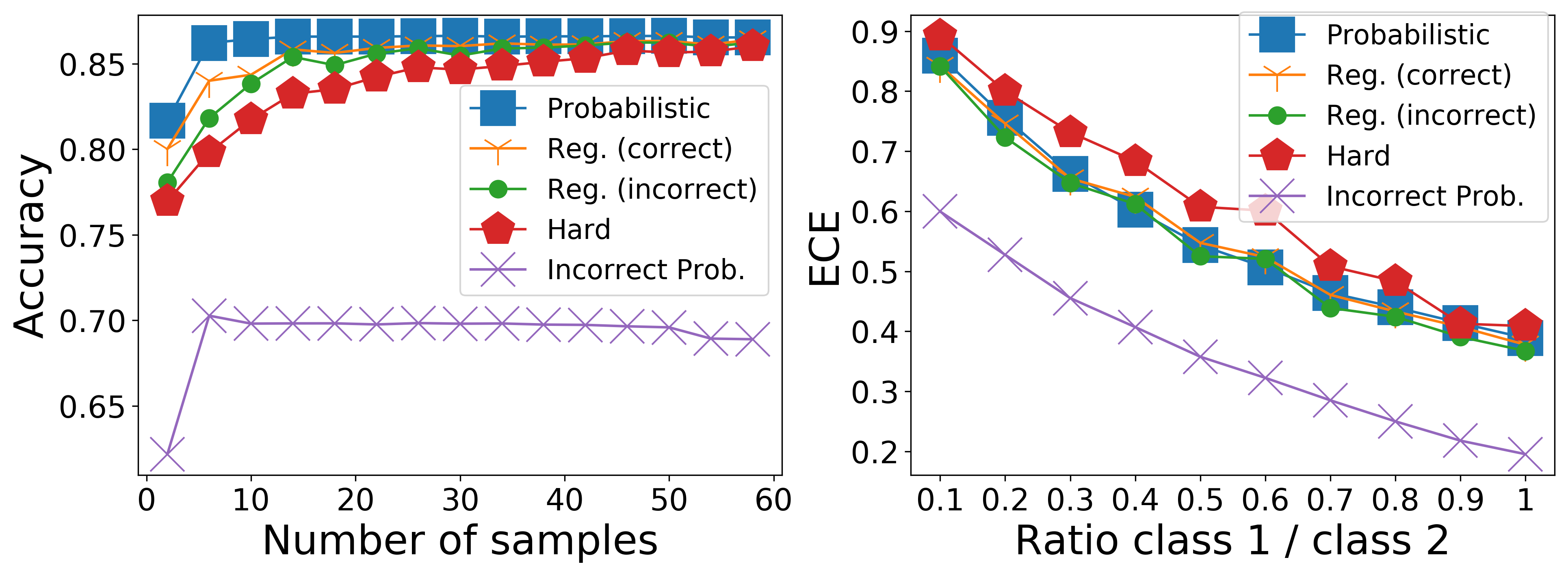} % Reduce the figure size so that it is slightly narrower than the column. Don't use precise values for figure width.This setup will avoid overfull boxes. 
\caption{Accuracy as a function of the number of training instances (left). Calibration as a function of the class imbalance (right).}
\label{fig:comparison}
\end{figure}

We trained models with different numbers of instances, starting with 2 (one from each Gaussian), and progressively increased the number until we used the 60 instances. We compared the accuracy of the models learned (respectively) with hard, correct and incorrect probabilistic labels, as well the regularized hard/probabilistic labels. Incorrect probabilistic labels are soft labels that do not represent true probabilities, but still are in [0,1], and all the entries add up to 1. We repeated these experiments 100 times, using logistic regression as the classifier, and show the average performance in
Figure~\ref{fig:comparison} (left). The second critical consideration is learning a calibrated classifier, a task whose complexity increases with unbalanced datasets. In a second experiment, we kept 10 training instances from Gaussian 1 and progressively changed the number of training instances of the second Gaussian from 1 to 10. Figure~\ref{fig:comparison} (right) shows the expected calibration error for different imbalance ratios.

Three things are important about this figure: 
(1)~ The probabilistic labels lead to more accurate classifiers when the number of instances is small. As this number increases, the model trained with hard labels can also learn the real probabilities. (2)~Providing incorrect probabilistic labels, and training the model using exclusively those labels, is worse than providing the hard labels. However, the regularized hard/probabilistic labels allow the model to converge to the real probabilities, even if the probabilistic labels are misleading. (3)~Probabilistic labels improve calibration, relative to hard labels, even in the presence of class imbalance.

% \begin{figure}
% \centering
% \includegraphics[width=0.7\columnwidth]{Images/AccVsSamples.png} % Reduce the figure size so that it is slightly narrower than the column. Don't use precise values for figure width.This setup will avoid overfull boxes. 
% \caption{Accuracy as a function of the number of training instances.}
% \label{fig:comparison}
% \end{figure}

Figure~\ref{fig:soft_hard} illustrates the change in the decision boundary changes when using probabilistic labels and a small training set. The boundary decision learnt by the hard labels tries to separate the 4 training instances perfectly (Figure~\ref{fig:soft_hard}, left); however, given the small training dataset the decision boundary does not generalize well 
(Figure~\ref{fig:soft_hard},~right). 
Probabilistic labels, on the hand, provide additional information that inform the learning algorithm ``how close'' each training instance is to the decision boundary. The decision boundary learnt by the probabilistic labels on Figure~\ref{fig:soft_hard}~(left) 
does not perfectly classify the training instances; however, it generalizes much better than the one learnt with hard labels. This simple experiment illustrates why we expect the probabilistic labels to generalize better when the number of training instances is small.

\subsection{Experiment 2: Hip Dysplasia}\label{sec:hip_experiment}
The machine learning task is to train a model that,
given an ultrasound image of the hip, identifies if it should be considered as normal or dysplastic. 
We used a private dataset collected from a multi-year clinical study of developmental dysplasia of the hip,
which contains 685 labeled, conventional, b-mode ultrasound images of the hip. A clinical expert labeled 342 of the images as dysplastic, while the remaining were considered normal. 

We divided this dataset into a training set (70 \% of the individuals = 429 images) and a hold-out set (30\% of the individuals = 256 images).  The hold-out set was used exclusively for testing purposes. After computing the soft labels using the procedure described in Section~\ref{sec:hip}, we trained a deep learning model that receives an image as an input and whose output is the probability of being diagnosed with dysplasia.

The deep neural network consisted of 5 layers (with 32, 64, 128, 256, and 512 3-by-3 convolutional filters). We added 2 fully connected layer with 1000 neurons in each hidden layer, and a single neuron in the output layer. The network used ReLU units as the activation function in the intermediate layers, and a sigmoid function in the output layer. The input to the network were ultrasound images resized to 128 by 128 pixels.

We compared four different training scenarios: with hard labels, soft labels, probabilistic labels, and regularized hard/probabilistic labels. For the soft labels experiments, we followed the approach proposed by Szegedy et al. (2016), setting $\epsilon = 0.1$~\cite{szegedy2016rethinking}  We evaluated the performance exclusively on the hold-out set measuring the classification accuracy, the area under the ROC curve, the HL-statistic, and the expected calibration error. Table~\ref{tab:results} shows the results.

\begin{table}[t]
\centering
\caption{Accuracy (threshold of 0.5 in the predictions) and area under the ROC curve (AUC) for performance (higher is better); and HL-Statistic and the expected calibration error (ECE) for calibration (lower is better)}\label{tab:results}
\begin{tabular}
{|>{\columncolor{LightCyan}}c|c c c c|}
\hline
\rowcolor{red!50} \textbf{Model} &  \multicolumn{4}{|c|}{\textbf{Dysplasia}}\\
\hline
\textbf{Labels} & \textbf{Hard} & \textbf{Soft} & \textbf{Prob} & \textbf{Reg}\\
\hline
\textbf{Accuracy} & 74\% & 68\% & 80\% & \textbf{83}\% \\
\textbf{AUC} & 0.82 & 0.74 & \textbf{0.87} & \textbf{0.87} \\
\textbf{HL Stat.} & $> 100$ & $59$ & $12.9$ &\textbf{9.2} \\
\textbf{ECE} & 0.56 & \textbf{0.34} & 0.41 & 0.46\\
% \hline
\end{tabular}
\begin{tabular}
{|>{\columncolor{LightCyan}}c|c c c c|}
\hline
\rowcolor{red!50} \textbf{Model} &  \multicolumn{4}{|c|}{\textbf{Fatty liver}}\\
\hline
\textbf{Labels} & \textbf{Hard} & \textbf{Soft} & \textbf{Prob} & \textbf{Reg}\\
\hline
\textbf{Accuracy} & \textbf{90}\% & 85\% & 81\% & \textbf{89}\% \\
\textbf{AUC} & \textbf{0.97} & 0.94 & 0.94 & \textbf{0.97} \\
\textbf{HL Stat.} & $> 100$ & 17 & \textbf{8.5} & 12.9 \\
\textbf{ECE} & 0.53 & \textbf{0.41} & \textbf{0.41} & 0.48\\
\hline
\end{tabular}
\begin{tabular}
{|>{\columncolor{LightCyan}}c|c c c c|}
\hline
\rowcolor{red!50} \textbf{Model} &  \multicolumn{4}{|c|}{\textbf{ Glaucoma }}\\
\hline
\textbf{Labels} & \textbf{Hard} & \textbf{Soft} & \textbf{Prob} & \textbf{Reg}\\
\hline
\textbf{Accuracy} & 55\% & 59\% & 69\% & \textbf{77}\% \\
\textbf{AUC} & 0.66 & 0.65 & 0.79 & \textbf{0.83} \\
\textbf{HL Stat.} & $> 100$ & $16.5$ &\textbf{12.8} &$15.3$ \\
\textbf{ECE} & 0.50 & 0.32 & \textbf{0.26} & 0.30\\
\hline
\end{tabular}
\end{table}

\subsection{Experiment 3: Fatty Liver}
\begin{figure}
\centering
\includegraphics[width=0.9\columnwidth]{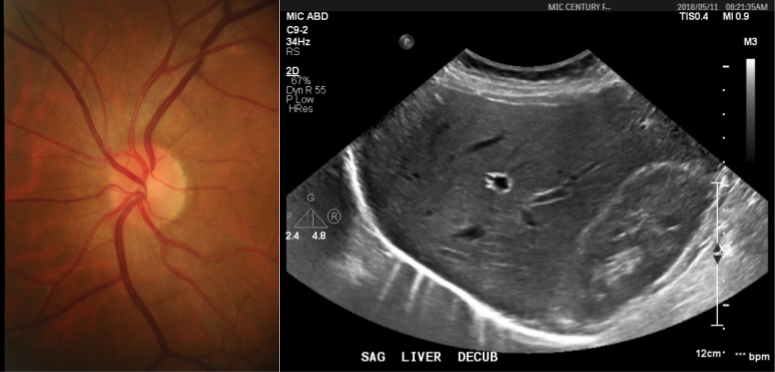} % Reduce the figure size so that it is slightly narrower than the column. Don't use precise values for figure width.This setup will avoid overfull boxes. 
\caption{Typical images used for the diagnosis of glaucoma (left) and fatty liver (right).}
\label{fig:glaucomaLiver}
\end{figure}

The machine learning task is to produce a model that predicts if an ultrasound image of the liver (see Figure~\ref{fig:glaucomaLiver}, right) should be diagnosed as fatty, or normal. We used a private dataset that contains 505 images, with labels made by an expert radiologist.
We placed 353 (70\%) in the training set, and the remaining 152 in the hold out set. The percentage of normal cases in both sets was 62\%.

Similarly to the experiment with hip dysplasia, we first trained a probabilistic model to encode the medical knowledge. Hamer et al. (2006) reports that the diagnosis of fatty liver depends on the difference in echogenicity among the liver, diaphragm, and the periportal zone~\cite{hamer2006fatty} --  more precisely, an increased hepatic echogenicity that obscures the periportal and diaphragm echogenicity suggest a liver is fatty~\cite{kim2005appropriateness}.

We manually defined regions of interest over the relevant anatomical parts in the images, and extracted their mean pixel intensity. The vector $Z=[m_1, m_2, m_3]$ contains the mean intensity value of the regions of interest extracted (respectively) from liver, diaphragm and periportal zone. This time, we modelled $P(Y\ |\ Z) = \sigma(Z^T\theta)$, where $\sigma(\cdot)$ is the sigmoid function. We then use the probabilistic labels $P(Y\ |\ Z)$ to train a network with the same architecture described in Section~\ref{sec:hip_experiment}, and perform the comparison among models trained with hard, soft, and probabilistic labels. Table~\ref{tab:results} shows the results.

% \begin{table}[t]
% \centering
% \caption{Results of the experiments for the fatty liver dataset. The HL statistic is used to measure the calibration of outputs (lower is better).}\label{tab:liver_results}
% \begin{tabular}{|>{\columncolor{LightCyan}}c|c c c c|}
% \hline
% \rowcolor{red!50} \textbf{Model} &  \multicolumn{4}{|c|}{\textbf{Fatty liver}}\\
% \hline
% \textbf{Labels} & Hard & Soft & Prob & Prior\\
% \hline
% \textbf{Accuracy} & 90\% & 85\% & 81\% & \textbf{89}\% \\
% \textbf{HL Stat.} & $> 100$ & $17$ & $8.5$ &$12.9$ \\
% \hline
% \end{tabular}
% \end{table}

\subsection{Experiment 4: Glaucoma}
Here, we wanted to learn a model that classifiers the fundus image of a retina (Figure~\ref{fig:glaucomaLiver}, left) as healthy, or as suspicious of glaucoma. We used the publicly available dataset 
RIM-ONE~r3~\cite{fumero2011rim}, which contains 85 images classified as normal, and 74 as suspects of glaucoma. We used 70\% of the data for training purposes, while the remaining 30\% was used as hold-out set. We kept the same healthy over glaucoma ratio in both datasets. Besides the diagnosis, which we used as ground truth, the dataset contains the masks of the disc and the cup masks of the optic nerve.

MacCormick et al. (2019) identifies the vertical and horizontal cup-to-disc-ratio 
as an important feature for the diagnosis of glaucoma~\cite{maccormick2019accurate}. The vector $Z=[r_1, r_2] \in \mathbb{R}^2$ contains these vertical and horizontal ratios. We assumed that the data comes from a mixture of 2 bi-variate Gaussians (one per each class) and, after learning the parameters of the probabilistic model, we used Eq.~\ref{eq:bayes} to compute the probabilistic labels.

We used the same network architecture than in previous experiments, and trained models with the same types of labels: hard, soft, and probabilistic. Table~\ref{tab:results} shows the results of the experiments. 

% \begin{table}[t]
% \centering
% \caption{Results of the experiments for the glaucoma dataset. The HL statistic is used to measure the calibration of outputs (lower is better).}\label{tab:glaucoma_results}
% \begin{tabular}{|>{\columncolor{LightCyan}}c|c c c c|}
% \hline
% \rowcolor{red!50} \textbf{Model} &  \multicolumn{4}{|c|}{\textbf{Glaucoma}}\\
% \hline
% \textbf{Labels} & Hard & Soft & Prob & Prior\\
% \hline
% \textbf{Accuracy} & 55\% & 59\% & 77\% & \textbf{83}\% \\
% \textbf{HL Stat.} & $> 100$ & $16.5$ & $12.8$ &$15.3$ \\
% \hline
% \end{tabular}
% \end{table}

\section{DISCUSSION}\label{sec:discussion}

Table~\ref{tab:results} (Dysplasia, Glaucoma) shows that the probabilistic labels, and the regularized hard/probabilistic approach greatly increases the accuracy of the classifiers when compared with the rest of the strategies. They also improve the calibration of the predicted outputs. It is important to highlight that all models were trained with the same architecture, and exactly the same training instances. The only change was the labels used during the training procedure. 

These results suggests that the network indeed benefits from labels that encode how close a given instance is to the decision boundary. On the other hand, the simple soft labels ($y \in \{0.1,0.9\}$), help to improve the calibration of the outputs; however, they do not improve the classification accuracy, suggesting that simply penalizing high confidence in the predictions is not enough to achieve good results.

\begin{figure}
\centering
\includegraphics[width=0.9\columnwidth]{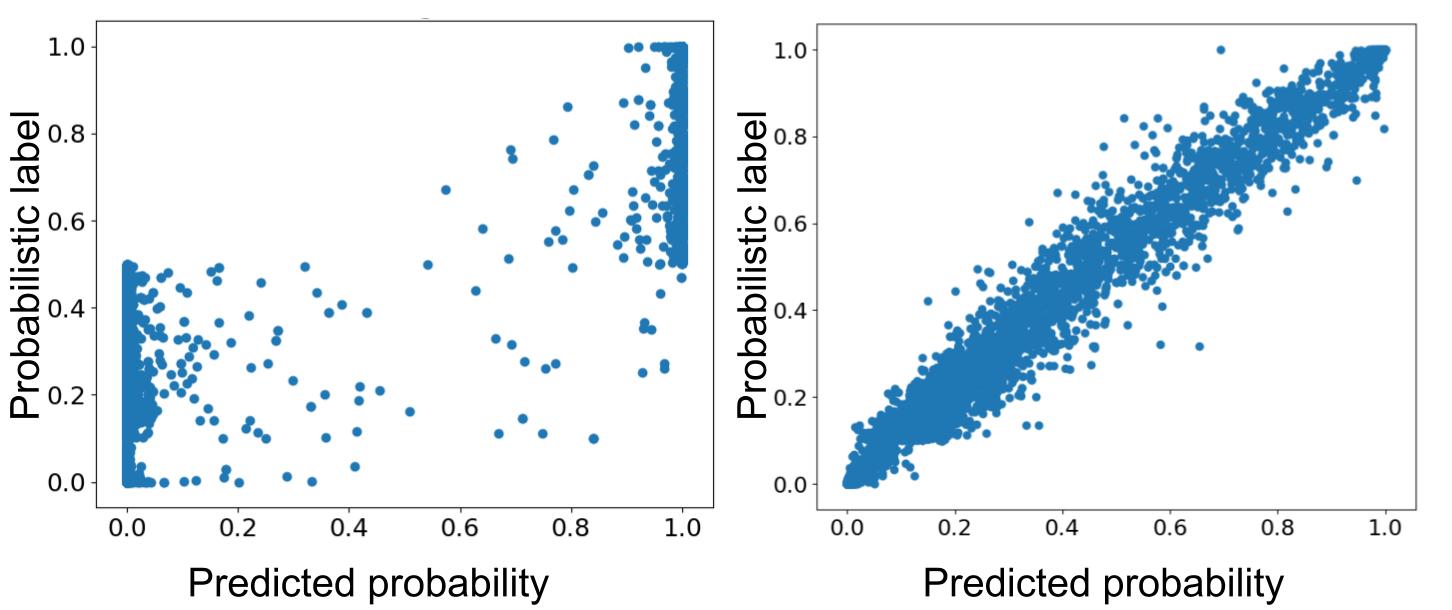} % Reduce the figure size so that it is slightly narrower than the column. Don't use precise values for figure width.This setup will avoid overfull boxes. 
\caption{Predicted values vs probabilities of a model trained with hard labels (left) and probabilistic labels (right) for the hip dysplasia experiment.}
\label{fig:Soft_Hard_Training}
\end{figure}

The liver dataset in Table~\ref{tab:results} shows an interesting case. The accuracy achieved by the model trained with hard labels is clearly superior to the one that used just probabilistic labels. As mentioned in Section~\ref{sec:prob_labels}, when the probabilistic labels are misleading, the model will converge to the wrong probabilities, which in turn increase the error in the predictions. This is not surprising, since the problem of diagnosis of fatty liver is very subjective in nature, and it is not clear what are the biomarkers that allow for an accurate diagnosis~\cite{strauss_interobserver_2007}. 

Despite the complexity of obtaining reliable probabilistic labels for the fatty liver task, the regularized model achieves almost the same performance as with the hard labels. This is a strong argument for the computation of probabilistic labels. The potential gains are important if they are reliable (9\% and 22\% in our experiment), while the potential losses are small (1\% in our experiments). Although computing the probabilistic labels require some manual measurements during training, the inference part is completely automated and does not require any manually extracted features.

Besides accuracy, for sensitive tasks, it is desirable to have models that not only make accurate predictions, but also that express their confidence on the predicted label in the form of probabilities. Our experiments indicate that the traditional training procedures for deep neural nets produce models whose real-valued output might not correspond to the probability of belonging to a specific class. Probabilistic labels are an alternative that not only improves the calibration of the outputs, but also improve the accuracy.

Figure~\ref{fig:Soft_Hard_Training} provides insight on why the model trained with the probabilistic labels is better calibrated. When using hard-labels, the deep learning models are penalized for not outputting 1 for the correct class, and 0 everywhere else. Naturally, their output tends to be closer to those values, so they tend to assign a ``high confidence'' to almost all their predictions. This is problematic for tasks where there is no clear boundary between the classes, such as medical diagnosis. Probabilistic labels, on the other side, encourage the model to make predictions in the entire range $[0, 1]$, assigning different degrees of confidence to the predictions.

The results from the different datasets strongly argue for the incorporation of probabilistic labels as priors when training classifiers. These labels allow for a more sample-efficient learning, since models using probabilistic labels provide equal or better accuracy than the ones trained with the traditional hard labels. Additionally the output of models that use probabilistic labels are better calibrated, providing a straightforward interpretation of the predictions as probabilities. The potential improvement in performance justifies the extra annotations needed during training when the size of the dataset is small, and when there is expert knowledge that can be leveraged during training. We anticipate that this approach will apply to any other tasks that maps high-dimensional inputs to categorical outputs, and where the features that determine the class can be encoded into a probabilistic model.
\acknowledgments{This research was partially supported by the Mexican National Council of Science and Technology (CONACYT), Canada's Natural Science and Engineering Research Council (NSERC) and the Alberta Machine Intelligence Institute (AMII). We also thank Medo.AI for providing the hip and liver datasets.}
\bibliography{aistats}
\bibliographystyle{abbrv}

% \begin{thebibliography}{}
% \setlength{\itemindent}{-\leftmargin}
% \makeatletter\renewcommand{\@biblabel}[1]{}\makeatother
% \bibitem{} J.~Alspector, B.~Gupta, and R.~B.~Allen (1989).
%     \newblock Performance of a stochastic learning microchip.
%     \newblock In D. S. Touretzky (ed.),
%     \textit{Advances in Neural Information Processing Systems 1}, 748--760.
%     San Mateo, Calif.: Morgan Kaufmann.

% \bibitem{} F.~Rosenblatt (1962).
%     \newblock \textit{Principles of Neurodynamics.}
%     \newblock Washington, D.C.: Spartan Books.

% \bibitem{} G.~Tesauro (1989).
%     \newblock Neurogammon wins computer Olympiad.
%     \newblock \textit{Neural Computation} \textbf{1}(3):321--323.
% \end{thebibliography}

\end{document}